\documentclass[sigconf,screen,nonacm,pbalance]{acmart}
\setcopyright{none}
\usepackage{mathtools}
\usepackage{algorithm}
\usepackage{algorithmic}
\usepackage{listings}
\usepackage{multirow}
\usepackage{makecell}
\usepackage{subcaption}
\usepackage[normalem]{ulem}
\usepackage{tikz}
\usepackage{fancyvrb}
\usepackage{framed}
\usepackage{array}
\usepackage{tabularx}
\usepackage{colortbl}
\usepackage{xspace}
\usepackage{placeins}
\usepackage[most]{tcolorbox}

\definecolor{questionborder}{RGB}{126,174,220}
\definecolor{questionbg}{RGB}{247,250,253}

\newtcolorbox{keyquestion}{
  width=\columnwidth,
  colback=questionbg,
  colframe=questionborder,
  boxrule=0.8pt,
  arc=2mm,
  left=5pt,
  right=5pt,
  top=4pt,
  bottom=4pt,
  before skip=6pt,
  after skip=6pt,
  fontupper=\itshape,
  enhanced
}

\definecolor{prompttitlegray}{RGB}{232,232,232}
\definecolor{rulegray}{RGB}{90,90,90}
\definecolor{caseblue}{RGB}{38,65,93}
\definecolor{casegray}{RGB}{247,248,250}
\definecolor{casegrayline}{RGB}{145,151,158}
\definecolor{casegreen}{RGB}{242,249,237}
\definecolor{casegreenline}{RGB}{174,207,151}
\definecolor{caseblueback}{RGB}{239,246,250}
\definecolor{promptnavy}{RGB}{30,40,60}
\definecolor{promptcontentbg}{RGB}{250,250,252}

\newcommand{\method}{\textsc{MemPrism}}

\newcolumntype{Y}{>{\raggedright\arraybackslash}X}
\newcolumntype{C}{>{\centering\arraybackslash}X}

\lstdefinestyle{PromptTextStyle}{%
  basicstyle=\ttfamily\footnotesize,
  breaklines=true,
  breakatwhitespace=false,
  breakautoindent=false,
  columns=flexible,
  keepspaces=true,
  xleftmargin=0pt,
  xrightmargin=0pt,
  aboveskip=0pt,
  belowskip=0pt
}
\newtcblisting[auto counter,number within=section]{promptbox}[1]{
  enhanced,
  breakable,
  colback=promptcontentbg,
  colframe=promptnavy,
  boxrule=1pt,
  arc=2mm,
  width=\linewidth,
  center,
  left=8pt, right=8pt, top=4pt, bottom=6pt,
  before skip=6pt, after skip=6pt,
  title={\small\bfseries\sffamily Prompt~\thetcbcounter: #1},
  colbacktitle=promptnavy,
  coltitle=white,
  fonttitle=\small\bfseries\sffamily,
  boxed title style={boxrule=0pt,colback=promptnavy,arc=2mm},
  attach boxed title to top left={xshift=0pt,yshift=0pt},
  listing only,
  listing options={style=PromptTextStyle}
}

\newcommand{\casebadge}[2]{%
  \tikz[baseline=(casebadge.base)]
    \node[draw=#1,fill=#1!10,text=#1,rounded corners=1.4pt,
      inner xsep=3.2pt,inner ysep=1.2pt,font=\sffamily\bfseries\scriptsize]
      (casebadge){#2};}
\newcommand{\casearrow}{%
  \par\centering\textcolor{caseblue!65}{\large$\downarrow$}%
  \par}

\newtcolorbox{caseflowbox}[1]{%
  enhanced,
  colback=white,
  colframe=caseblue,
  boxrule=0.9pt,
  arc=3mm,
  left=3.5mm,right=3.5mm,top=2.2mm,bottom=2.2mm,
  title={#1},
  colbacktitle=white,
  coltitle=caseblue,
  fonttitle=\large\bfseries,
  boxed title style={boxrule=0pt,colback=white},
  attach boxed title to top left={xshift=2mm,yshift=-1.4mm},
  before skip=2mm,after skip=1mm}

\newtcolorbox{casepromptbox}[1]{%
  enhanced,
  colback=casegray,
  colframe=casegrayline,
  boxrule=0.55pt,
  arc=2mm,
  left=2.5mm,right=2.5mm,top=1.5mm,bottom=1.5mm,
  title={#1},
  colbacktitle=casegray,
  coltitle=black,
  fonttitle=\bfseries\small,
  before skip=1.5mm,after skip=1.5mm}

\newtcolorbox{casevisualbox}[1]{%
  enhanced,
  colback=white,
  colframe=casegrayline,
  boxrule=0.55pt,
  arc=2mm,
  left=2.5mm,right=2.5mm,top=1.5mm,bottom=1.5mm,
  title={#1},
  colbacktitle=white,
  coltitle=black,
  fonttitle=\bfseries\small,
  before skip=1.5mm,after skip=1.5mm}

\newtcolorbox{caseresponsebox}[1]{%
  enhanced,
  colback=casegreen,
  colframe=casegreenline,
  boxrule=0.65pt,
  arc=2.5mm,
  left=3mm,right=3mm,top=1.7mm,bottom=1.7mm,
  title={#1},
  colbacktitle=casegreen,
  coltitle=black,
  fonttitle=\bfseries\small,
  before skip=1.5mm,after skip=1.5mm}

\newtcolorbox{caseresultbox}[1]{%
  enhanced,
  colback=caseblueback,
  colframe=caseblue!55,
  boxrule=0.65pt,
  arc=2.5mm,
  left=3mm,right=3mm,top=1.7mm,bottom=1.7mm,
  title={#1},
  colbacktitle=caseblueback,
  coltitle=black,
  fonttitle=\bfseries\small,
  before skip=1.5mm,after skip=1.5mm}

\title[MemPrism: Task-Conditioned Relational Memory Views]{MemPrism: Task-Conditioned Relational Memory Views for Long-Horizon Agents}
\makeatletter
\let\MemPrismDefaultFootnoteSymbol\@fnsymbol
\newcommand{\MemPrismAuthorFootnoteSymbol}[1]{%
  \ifnum#1=1\relax
    \scalebox{2}{\MemPrismDefaultFootnoteSymbol{#1}}%
  \else
    \MemPrismDefaultFootnoteSymbol{#1}%
  \fi}
\AddToHook{cmd/maketitle/before}{\let\@fnsymbol\MemPrismAuthorFootnoteSymbol}
\AddToHook{cmd/maketitle/after}{\let\@fnsymbol\MemPrismDefaultFootnoteSymbol}
\makeatother

\author{Bingfan Zeng}
\authornote{Equal contribution. Contact: \href{mailto:zhisheng.researcher@gmail.com}{\nolinkurl{zhisheng.researcher@gmail.com}}.}
\affiliation{
  \institution{South China University of Technology}
  \city{Guangzhou}
  \country{China}
}

\author{Zhisheng Chen}
\authornotemark[1]
\affiliation{
  \institution{Nanyang Technological University}
  \city{Singapore}
  \country{Singapore}
}

\author{Zhengwei Xie}
\authornotemark[1]
\affiliation{
  \institution{University of Science and Technology of China}
  \city{Hefei}
  \country{China}
}

\author{Bangde Cao}
\affiliation{
  \institution{Beijing University of Posts and Telecommunications}
  \city{Beijing}
  \country{China}
}

\author{Yuxuan Li}
\affiliation{
  \institution{Nanyang Technological University}
  \city{Singapore}
  \country{Singapore}
}

\author{Jinhan Li}
\affiliation{
  \institution{University of Science and Technology of China}
  \city{Hefei}
  \country{China}
}

\author{Zheng Lu}
\affiliation{
  \institution{Peking University}
  \city{Beijing}
  \country{China}
}

\author{Xiangchen Guan}
\affiliation{
  \institution{Peking University}
  \city{Beijing}
  \country{China}
}

\author{Zikai Xiao}
\affiliation{
  \institution{Zhejiang University}
  \city{Hangzhou}
  \country{China}
}

\author{Rui Qian}
\authornote{Corresponding authors.}
\affiliation{
  \institution{Fudan University}
  \city{Shanghai}
  \country{China}
}

\author{Jingwei Song}
\authornotemark[2]
\affiliation{
  \institution{Shanghai Jiao Tong University}
  \city{Shanghai}
  \country{China}
}

\renewcommand{\shortauthors}{Zeng et al.}

\begin{document}

\begin{abstract}
Long-horizon agents performing multi-step Web tasks must reuse earlier page observations and interactions to support subsequent actions. However, preserving relevant history does not ensure that it exposes the relations needed for the current decision. We call this gap \textit{\underline{representation mismatch}}. To this end, we propose \textit{\textbf{MemPrism}}, a task-conditioned relational memory framework that separates persistent experience storage from decision-time working memory. MemPrism records interactions as the event stream and dynamically constructs relational views according to the current task context. A lightweight view policy selects the relation structure, evidence range, outcome condition, and granularity, while a deterministic composer and renderer transform historical facts into a temporary optical working-memory view for a frozen task policy. Experiments on long-horizon embodied and web-agent benchmarks show that MemPrism consistently improves the task performance, especially as trajectories become longer, while reducing memory token consumption. Furthermore, the learned view policy transfers across different VLMs without additional adaptation, demonstrating the effectiveness of task-conditioned relational views as a general memory interface for agents.
\par\noindent\textbf{Code:} \url{https://github.com/Feld-maxiu/MemPrism}.
\end{abstract}

\ccsdesc[500]{Computing methodologies~Artificial intelligence}
\ccsdesc[300]{Information systems~World Wide Web}
\keywords{Agent memory, long-horizon agents, relational memory views, optical memory, web agents}

\maketitle
\hypersetup{pdfauthor={Bingfan Zeng, Zhisheng Chen, Zhengwei Xie, Bangde Cao, Yuxuan Li, Jinhan Li, Zheng Lu, Xiangchen Guan, Zikai Xiao, Rui Qian, Jingwei Song}}


\section{Introduction}
\label{sec:introduction}

For long-horizon agents, past experience becomes useful memory only when it is organized to support the current decision. This challenge is becoming increasingly important as Large language model (LLM) and vision-language model (VLM) based agents tackle web navigation~\cite{zhou2024webarena}, software engineering~\cite{yang2024swe}, tool use~\cite{qin2023toolllm}, embodied interaction, and open-world exploration~\cite{wang2023voyager}, where trajectories often span tens or even hundreds of steps. At this scale, success depends not only on local reasoning, but also on how effectively the agent can structure and use its growing interaction history. For Web agents, later actions may depend on inputs, selections, and intermediate results from earlier pages, making interaction history an important source of decision context.

\begin{figure}[t!]
\centering
\includegraphics[width=\linewidth]{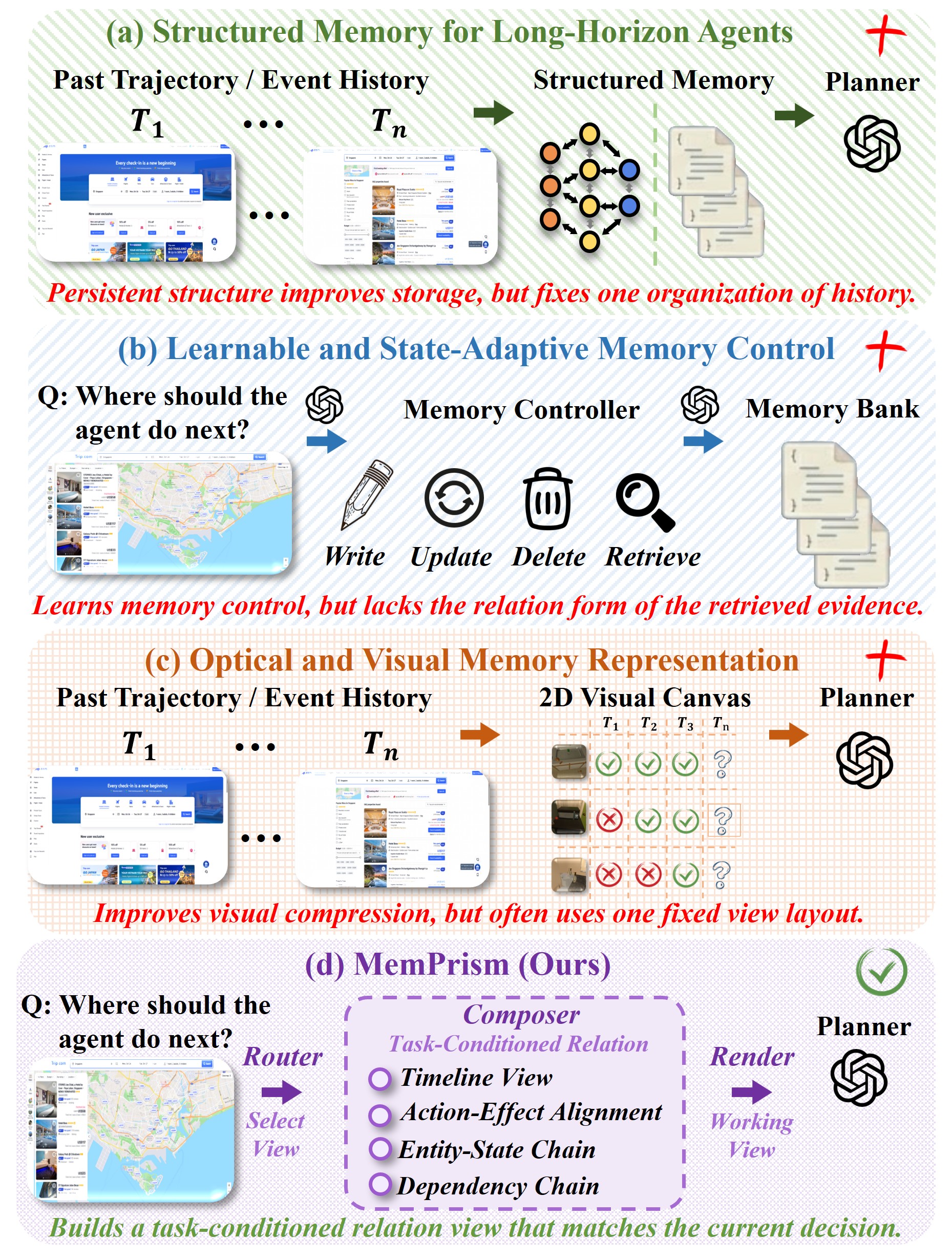}
\caption{Conceptual comparison of different memory paradigms for long-horizon
agents. Existing approaches improve memory organization, control, or
representation medium, but generally provide history in a fixed form
for downstream decisions. In contrast, MemPrism preserves stable
interaction facts as an event stream and dynamically constructs
task-conditioned relational views, allowing the same history to support different decision requirements.}
\Description{Comparison of fixed memory representations with task-conditioned relational views constructed from a persistent event stream.}
\label{fig:Difference}
\end{figure}

Recent work has advanced agent memory along three complementary directions. One line organizes raw trajectories into persistent structures, such as hierarchical summaries~\cite{lee2024human}, knowledge graphs~\cite{anokhin2024arigraph}, and fact stores~\cite{NEURIPS2025_19909c36}. Another makes memory control itself learnable by treating writing, updating, deletion, and retrieval as part of the agent policy~\cite{yu2026agentic}. A third revisits the representation medium, optical memory renders long interaction histories as compact images to reduce token cost~\cite{feng-etal-2026-agentocr}. These directions improve how memory is stored, managed, and accessed. However, they often retain a common assumption: once relevant evidence has been preserved and retrieved, it can be delivered to the policy through a fixed representation~\cite{yao2022react,park2023generative,wang2023augmenting}.

However, a unique representation is not equally useful for every decision~\cite{zeng2024structural}. The same history may need to be organized in different ways for different tasks. Loop detection benefits from aligning repeated actions with their outcomes. State tracking requires changes to the same entity to be grouped over time, such as grouping changes to a Web form's field values and selected options throughout a workflow. Failure analysis links actions to environment feedback, while dependency analysis connects related events that may be far apart in the trajectory. The timeline, state table, action-effect view, and dependency trace may all be built from the same facts, yet each highlights a different type of evidence.

As a result, an agent may still fail even when the correct evidence has been stored and retrieved~\cite{pmlr-v306-he26am}. We call this failure \emph{representation mismatch}. It occurs when the available evidence is not organized in the relation form needed by the current subtask. This failure is different from storage failure and retrieval failure. It lies at the read-time interface between retrieved evidence and the task policy and also reveals a key observation in long-horizon memory:

\begin{keyquestion}
\textbf{Does access to the right evidence guarantee that it will be useful for the current decision?}
\end{keyquestion}

This observation suggests a principle: persistent memory should preserve stable and reusable interaction facts~\cite{hu2025hiagent}, while working memory should be constructed for the current decision. Rather than relying on a fixed representation, the agent should reorganize the same history according to the current observation, task goal, and execution state. Therefore, effective memory requires learning not only what to store and retrieve but how to present retrieved evidence to the policy.

To this end, we propose \textbf{MemPrism}, a task-conditioned relational working-memory framework. MemPrism records real interactions in a unified event stream. At each step, a lightweight view policy selects the relation type, evidence range, outcome filter, and level of detail. A deterministic pipeline then constructs a task-conditioned view from the event stream, which a frozen VLM uses to produce the next action. This view is temporary and serves only the current decision; it is never written back to persistent memory. Only executed actions and the resulting environment feedback are appended to the event stream.

MemPrism uses optical views as a unified output format, not only to compress history but also to make relations explicit. Its two-dimensional layout encodes structure directly: spatial proximity groups related actions, row and column alignment supports state comparison, arrows indicate change or dependency, and highlights draw attention to failures and task-relevant evidence. Thus, optical memory serves not merely as a compact representation but as a structured interface for task-conditioned relations.

Our main contributions are as follows:
\begin{itemize}
    \item \textbf{We formulate post-retrieval representation mismatch.}
    We distinguish this failure from missing storage or failed retrieval:
    the required evidence is available, but its representation does not
    expose the relations needed for the current decision. We therefore
    formulate the construction of task-specific relational views as an
    explicit working-memory decision problem.

    \item \textbf{We decouple persistent history from temporary working views.}
    MemPrism stores interactions in a unified and reusable event
    stream, while constructing a new working view on demand at each
    decision step. This separation avoids irreversible distortion
    caused by fixed summaries, repeated compression, or previously
    generated views.

    \item \textbf{We introduce a relational optical view space.}
    Each view is jointly defined by its relation type, evidence range,
    outcome condition, and level of detail. Two-dimensional
    layouts explicitly represent temporal order, action-outcome relations,
    entity-state changes, and local dependencies. 

    \item \textbf{We introduce a task-conditioned view policy.}
    We initialize the policy with action-conditioned soft supervision and
    further optimize it with trajectory-level grouped GRPO, which isolates the effect of view selection and allows performance
    gains to be more directly attributed to the memory representation
    interface.
\end{itemize}


\section{Related Work}
\label{sec:related-work}

\subsection{Memory Organization and Retrieval}
\label{sec:rw-structured-memory}

HiAgent~\citep{hu2025hiagent} organizes interaction histories around subgoals.
AriGraph~\citep{anokhin2024arigraph} combines episodic and semantic memory in a graph, while A-MEM~\citep{NEURIPS2025_19909c36} builds an evolving memory network.
MAGMA further introduces multiple relational graphs to support intent-guided retrieval through adaptive graph traversal~\citep{jiang2026magma}.
Despite these advances, existing methods primarily focus on storing and retrieving experiences, giving less attention to how the same evidence should be organized to support different decisions.
\textit{On the Structural Memory of LLM Agents}~\citep{zeng2024structural} finds that the benefits of different memory structures vary across tasks, while LongMemEval~\citep{wu2024longmemeval} and ATM-Bench~\citep{mei2026according} show that reading strategies and evidence formats affect performance even with oracle evidence.
Building upon these insights, MemPrism shifts the focus from \textbf{retrieving relational evidence to learning its decision-time organization}, dynamically constructing task-conditioned relational views from persistent interaction histories.
This enables the same evidence to be reorganized into different working-memory structures according to the reasoning demands of each action.

\begin{figure*}[!t]
    \centering
    \includegraphics[width=\textwidth]{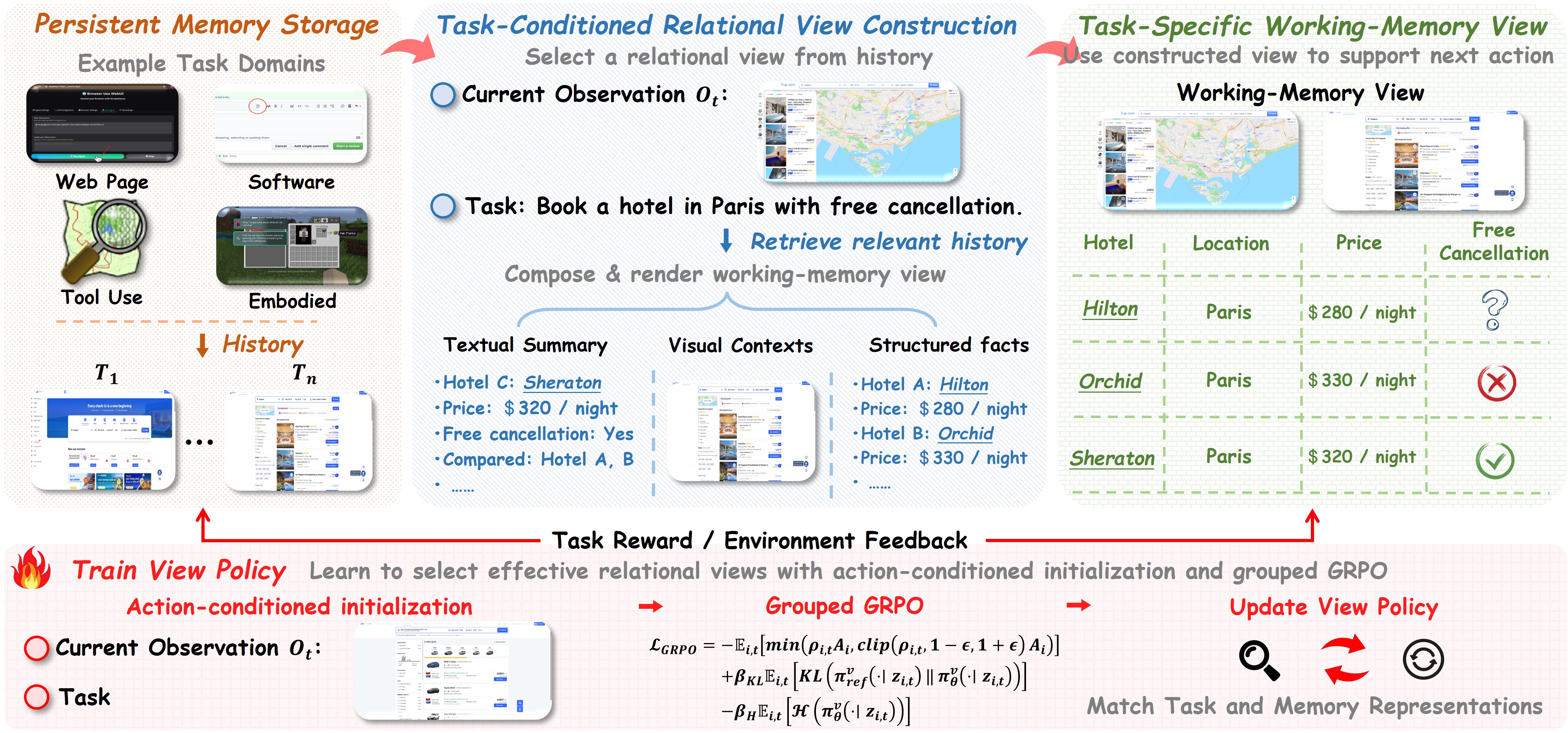}
    \caption{Overview of MemPrism. A learned view policy selects a relational view specification for the current decision. The view constructor organizes historical evidence and renders a temporary visual working-memory view, which a frozen task policy uses to generate an environment action. The view policy is initialized with action-conditioned teacher preferences and further optimized with grouped GRPO, while the task policy and view-construction modules remain fixed.}
    \Description{MemPrism organizes persistent interaction history into a temporary relational view for a frozen task policy. Only the view policy is trained, using action-conditioned supervision and grouped GRPO.}
    \label{fig:MemPrism_overall}
\end{figure*}

\subsection{Learnable Memory Policies}
\label{sec:rw-memory-control}

Recent work increasingly treats memory utilization as a learnable decision process.
Memory-R1~\citep{yan2026memory} optimizes memory writing, updating, deletion, and retrieval through reinforcement learning, while AgeMem~\citep{yu2026agentic} integrates long- and short-term memory operations into the agent's action space.
MRAgent~\citep{ji2026memory} further learns state-adaptive access to distant interaction histories through compact memory cues.
More closely related, MemCompiler~\citep{ding2026memcompiler} learns to compile state-relevant experience into textual and latent guidance for a frozen executor.
In contrast, MemPrism learns a task-conditioned policy over relational memory views, optimizing how historical evidence is organized and presented for each decision.

\subsection{Optical and Visual Memory Representation}
\label{sec:rw-optical-memory}

Optical memory converts long text histories into two-dimensional images~\citep{li2026ocr}.
AgentOCR renders observation-action traces as compact images and learns a dynamic compression rate~\citep{feng-etal-2026-agentocr}.
MemOCR further assigns visual space based on layout and information importance~\citep{shi2026memocr}.
These methods focus on storing or compressing history more efficiently.
MemPrism instead uses two-dimensional layout to express relation semantics. Spatial proximity, alignment, arrows, and highlights expose action effects, entity states, temporal change, and local dependencies.
Thus, optical memory in MemPrism is not only a compact medium. It is the output space of a task-conditioned relational memory interface.

\section{Method}
\label{sec:method}

\subsection{Task-Conditioned Relational Memory}
\label{sec:relational-memory}

As shown in Figure~\ref{fig:MemPrism_overall}, MemPrism decouples persistent memory storage from read-time organization. It formulates each interaction step as two successive decisions: how to organize historical evidence and which environment action to take.

Consider an agent with task goal $g$. At step $t$, it receives observation $o_t$ and maintains the history of completed interactions,
\begin{equation}
\mathcal{H}_{<t}=(h_1,\ldots,h_{t-1}),
\label{eq:interaction-history}
\end{equation}
where $h_i=(o_i^{-},u_i,y_i^{+},m_i)$ records the pre-action observation, executed action, environmental feedback, and associated metadata.

Given the current decision context, a learned view policy $\pi_\theta^v$ selects a view action $a_t^v$. The view constructor $\mathcal{C}$ then builds a temporary working-memory view from the history:
\begin{equation}
V_t=\mathcal{C}(\mathcal{H}_{<t},g,a_t^v).
\label{eq:working-view}
\end{equation}
The frozen task policy generates the next environment action:
\begin{equation}
u_t\sim\pi_{\mathrm{task}}
\left(\cdot\mid X_t,g,H_t^{\mathrm{short}},V_t\right),
\label{eq:task-policy}
\end{equation}
where $X_t$ denotes the current task input and $H_t^{\mathrm{short}}$ denotes the shared short interaction context.

After execution, the resulting interaction is appended to $\mathcal{H}$, while the temporary view $V_t$ is discarded and never written back to the persistent history.

\subsection{Factorized Relational View Space}
\label{sec:relational-view-space}

Choosing a relation type alone does not fully specify a memory view. The agent must also decide how far back to look, which events to include based on their outcomes, and how much detail to present. We therefore define a factorized relational view space in which these choices are made jointly for the current decision.

Each view is specified by
\begin{equation}
a_t^v=(\tau_t,w_t,c_t,\gamma_t)\in\mathcal{A}^v,
\label{eq:view-action}
\end{equation}
where $\mathcal{A}^v=\mathcal{T}\times\mathcal{W}\times\mathcal{O}\times\mathcal{G}$.
The four factors control the relational structure $\tau$, temporal range $w$, outcome condition $c$, and representation granularity $\gamma$, respectively.

\paragraph{Relational structure.}
Interaction histories provide temporal order, actions, environmental feedback, and information about entities and state changes. We use these signals to construct four complementary relational views.

\begin{figure*}[!t]
    \centering
    \includegraphics[width=0.93\textwidth]{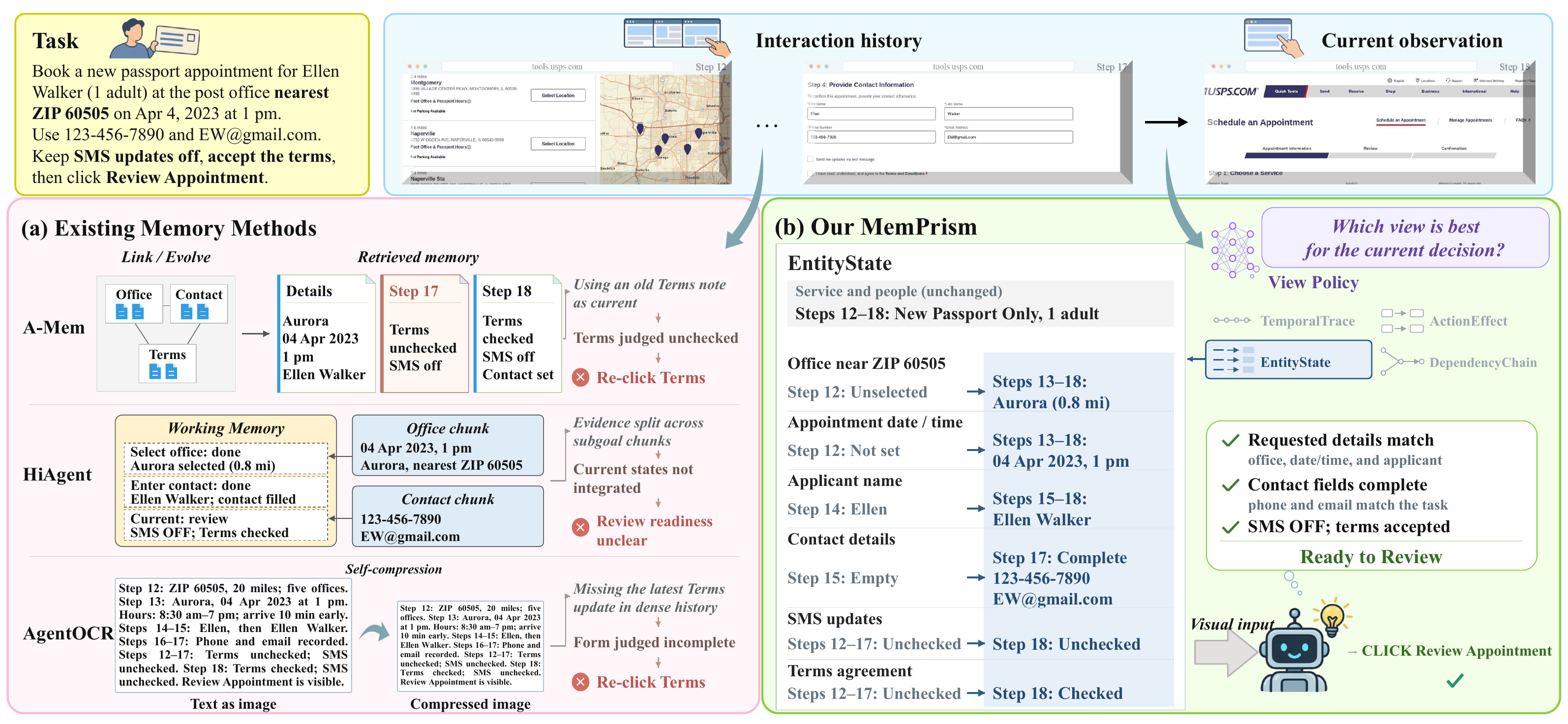}
    \caption{A Mind2Web case study. (a) Memory representations and decision errors of existing methods. (b) MemPrism's EntityState view aligns field states and supports the correct next action: click Review Appointment.}
    \Description{A web appointment task compares A-Mem, HiAgent, AgentOCR, and MemPrism using actual experimental data. A-Mem acts on an outdated Terms state; HiAgent leaves review readiness unresolved across subgoal chunks; AgentOCR misses the latest Terms update in its chronological history image. MemPrism aligns field states, showing that the terms are accepted and SMS updates remain off, and correctly selects Review Appointment.}
    \label{fig:web_case}
\end{figure*}

\begin{itemize}
\item \textbf{TemporalTrace.} This view presents events in the order they occurred, preserving their temporal sequence to show the interaction process and task progress.

\item \textbf{ActionEffect.} This view groups events by action and presents their observed outcomes and state changes, enabling comparisons across events with the same action.

\item \textbf{EntityState.} This view organizes state updates by entity and state key, bringing successive updates together to track how the same entity's state changes over time. Figure~\ref{fig:web_case} illustrates this organization in a Web form.

\item \textbf{DependencyChain.} This view uses the latest event as an anchor and links relevant preceding events through shared entities, state keys, and temporal proximity, organizing these interactions to provide local dependency cues.
\end{itemize}

\paragraph{Evidence selection.}
Even an appropriate relational structure may remain uninformative if necessary interactions are omitted or irrelevant events dominate the view. The temporal-scope factor $w$ determines history coverage through three scopes: \textbf{Recent-Short} retains recent local interactions; \textbf{Recent-Long} retains a broader range of recent interactions; and \textbf{All-History} retains the complete interaction history. These scopes span local, extended recent, and full-history contexts, allowing the view policy to adapt the temporal coverage of evidence to the current decision. The outcome condition $c$ further selects among all outcomes, exceptions, state updates, and interactions with no observed change.

\paragraph{Representation granularity.}
Even with an appropriate relational structure and evidence scope, presenting every interaction in full detail may obscure the information most relevant to the current decision. The granularity factor $\gamma$ controls the amount of detail retained through coarse, medium, and fine configurations. Coarser views emphasize representative events and relational changes, whereas finer views preserve more individual interactions and their details.

\paragraph{Joint view space.}
Together, these factors form the view space. Their Cartesian product allows the view policy to select all four factors jointly, capturing interactions among the design choices.

\subsection{Task-Conditioned View Policy}
\label{sec:view-routing}

MemPrism employs a lightweight view policy $\pi_\theta^v$ that selects a relational view specification conditioned on the current observation, task goal, and interaction history.

To support context-aware view selection without encoding the entire interaction history, we construct a compact, multi-scale representation of the interaction state. Specifically, we derive complementary local and global descriptors:
\begin{equation}
\mathbf{d}_t^{\mathrm{loc}}=\psi_{\mathrm{loc}}(\mathcal{H}_{<t}^{(K)}),\qquad
\mathbf{d}_t^{\mathrm{glob}}=\psi_{\mathrm{glob}}(\mathcal{H}_{<t}^{(M)}).
\label{eq:history-statistics}
\end{equation}
The local descriptor captures short-term action--feedback dynamics, while the global descriptor summarizes broader trajectory characteristics. Together, they provide lightweight temporal context for selecting decision-relevant relational views.

For view selection, a frozen encoder $e$ encodes the observation representation and the task goal. The resulting embeddings and statistical features are then projected into a shared space. A lightweight Transformer aggregates these inputs to obtain a decision representation,
\begin{equation}
z_t=\operatorname{Enc}_{\theta}
\left(e(o_t),e(g),\mathbf{d}_t^{\mathrm{loc}},\mathbf{d}_t^{\mathrm{glob}}\right).
\label{eq:router-state}
\end{equation}
The view policy uses a single joint prediction head to produce a probability distribution over the full view space:
\begin{equation}
\pi_\theta^v(a\mid z_t)
=\operatorname{softmax}(f_\theta(z_t))_a,
\qquad a\in\mathcal{A}^v.
\label{eq:view-policy}
\end{equation}
The joint prediction head models combinations of the four view factors rather than predicting each factor independently. View actions are sampled from this distribution during training; at inference time, the specification with the highest probability is selected.

\begingroup
\definecolor{mainrule}{gray}{0.5}
\definecolor{mainheader}{HTML}{E2E2E2}
\definecolor{maingroup}{HTML}{EFEFEF}
\definecolor{mainmethod}{HTML}{EAF4FF}
\begin{table*}[t]
\centering
\caption{Results on \textbf{Qwen2.5-VL-7B-Instruct} and \textbf{Qwen3.5-4B}. Values are action accuracies (\%) on Mind2Web, and success rates (\%) on ALFWorld and EB-ALFRED. Web., Dom., Com., and Cplx.\ denote Website, Domain, Common, and Complex, respectively.}
\label{tab:main_results}
\normalfont\normalsize
\setlength{\tabcolsep}{1pt}
\renewcommand{\arraystretch}{1.12}
\newcommand{\mainqwenmodel}[1]{%
  \raisebox{-0.10em}[0pt][0pt]{\includegraphics[height=0.85em,trim=23.94bp 22.86bp 241.20bp 28.17bp,clip]{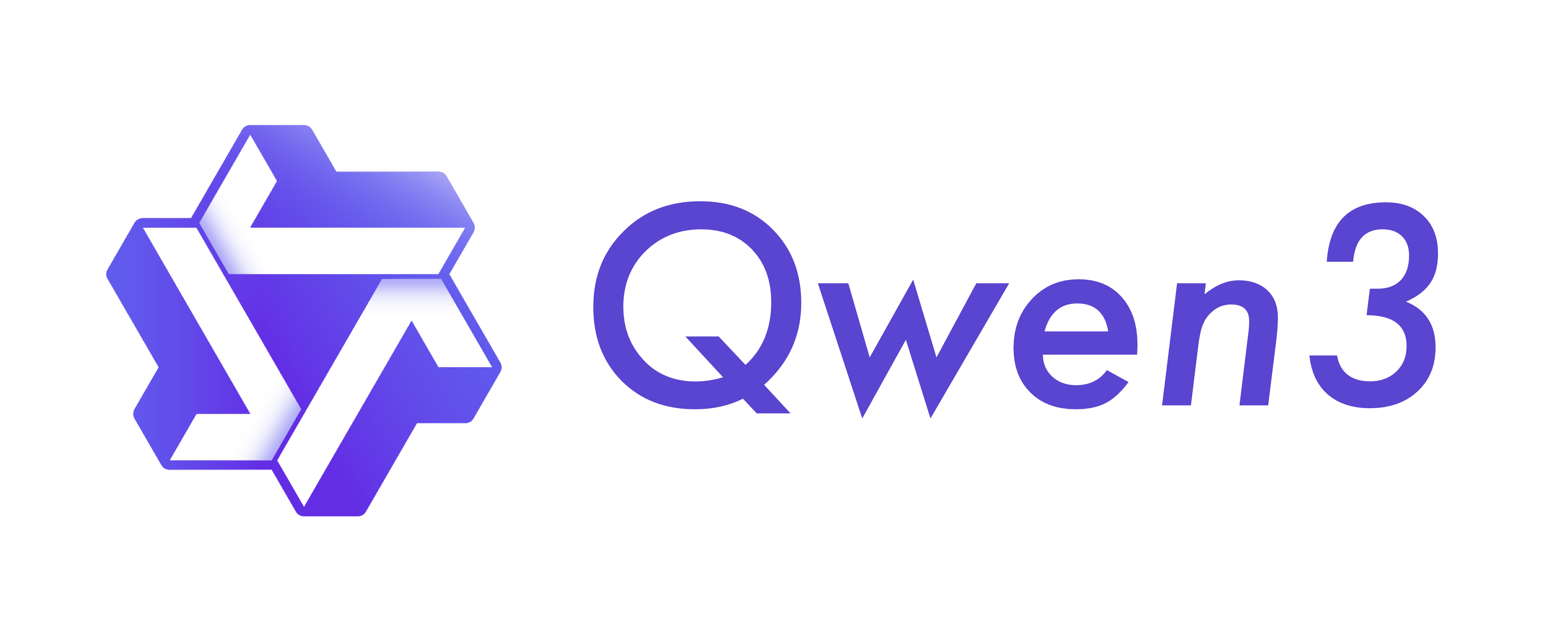}}%
  \hspace{0.35em}\textit{#1}}
\newcommand{\maintablecontents}{%
\arrayrulecolor{black}\toprule[0.7pt]
& \multicolumn{4}{c}{\textbf{Mind2Web}} & \multicolumn{7}{c}{\textbf{ALFWorld}} & \multicolumn{5}{c}{\textbf{EB-ALFRED}} \\
\arrayrulecolor{mainrule}\cmidrule(lr){2-5}\cmidrule(lr){6-12}\cmidrule(lr){13-17}
\rowcolor{mainheader}
\textbf{Method} & Task & Web. & Dom. & Overall & \shortstack{Pick \&\\Place} & Look & Clean & Heat & Cool & \shortstack{Pick2 \&\\Place} & Avg. & Base & Com. & Cplx. & Long & Avg. \\
\midrule[0.3pt]
\rowcolor{maingroup}
\multicolumn{17}{c}{\mainqwenmodel{Qwen2.5-VL-7B-Instruct}} \\
\cmidrule(lr){1-17}
No Memory & 12.6 & 8.0 & 8.0 & 8.9 & 67.6 & 35.4 & 19.3 & 31.3 & 30.1 & 4.4 & 33.6 & 16.0 & 10.0 & 14.0 & 12.0 & 13.0 \\
AgentOCR & 10.4 & 7.1 & 8.6 & 8.7 & 61.0 & 33.2 & 17.2 & 11.6 & 12.5 & 16.5 & 28.0 & 10.0 & 12.0 & 8.0 & 12.0 & 10.5 \\
\cmidrule(lr){1-17}
LangMem & 13.2 & 8.9 & 8.4 & 9.4 & 69.3 & 39.3 & 23.2 & 26.4 & 26.3 & 15.3 & 35.8 & 16.0 & 16.0 & 16.0 & 14.0 & 15.5 \\
A-Mem & 10.9 & 7.6 & 8.1 & 8.6 & 70.7 & 40.7 & 24.6 & 27.8 & 27.1 & 16.5 & 37.0 & 8.0 & 10.0 & 12.0 & 6.0 & 9.0 \\
Mem0 & 11.4 & 7.4 & 7.3 & 8.1 & 56.8 & 57.7 & 31.3 & 40.3 & 25.6 & 13.3 & 37.1 & 16.0 & 16.0 & 14.0 & 12.0 & 14.5 \\
Mem0\textsuperscript{g} & 11.1 & 7.1 & 7.1 & 7.9 & 61.9 & 44.0 & 25.3 & 37.8 & 31.6 & 11.2 & 36.3 & 14.0 & 16.0 & 16.0 & 12.0 & 14.5 \\
SimpleMem & 12.0 & 7.7 & 7.6 & 8.4 & 64.3 & 33.2 & 19.9 & 12.5 & 33.2 & 3.8 & 31.0 & 10.0 & 8.0 & 10.0 & 8.0 & 9.0 \\
MemoryOS & 12.8 & 8.4 & 7.5 & 8.6 & 63.2 & 45.9 & 27.9 & 39.9 & 34.0 & 14.2 & 38.5 & 16.0 & 12.0 & 14.0 & 14.0 & 14.0 \\
LightMem & 12.0 & 7.7 & 8.3 & 8.9 & 67.5 & 37.2 & 23.2 & 25.3 & 34.7 & 8.3 & 35.3 & 12.0 & 12.0 & 12.0 & 10.0 & 11.5 \\
\cmidrule(lr){1-17}
\rowcolor{mainmethod}
\textbf{MemPrism}\,\textnormal{\fontsize{8}{9.5}\selectfont (SFT)} & 13.0 & 10.1 & 10.9 & \textbf{11.2} & 56.6 & 46.9 & 18.9 & 41.6 & 31.5 & 10.3 & \textbf{34.3} & 20.0 & 20.0 & 22.0 & 16.0 & \textbf{19.5} \\
\rowcolor{mainmethod}
\textbf{MemPrism}\,\textnormal{\fontsize{8}{9.5}\selectfont (SFT+GRPO)} & 14.2 & 11.1 & 12.8 & \textbf{12.8} & 62.0 & 52.1 & 25.8 & 48.9 & 38.6 & 20.9 & \textbf{41.4} & 24.0 & 22.0 & 26.0 & 18.0 & \textbf{22.5} \\
\midrule[0.5pt]
\rowcolor{maingroup}
\multicolumn{17}{c}{\mainqwenmodel{Qwen3.5-4B}} \\
\cmidrule(lr){1-17}
No Memory & 14.1 & 8.6 & 8.4 & 9.5 & 70.0 & 46.2 & 40.7 & 6.2 & 8.0 & 54.2 & 41.1 & 18.0 & 10.0 & 18.0 & 10.0 & 14.0 \\
AgentOCR & 14.9 & 9.3 & 9.2 & 10.3 & 72.8 & 54.6 & 43.3 & 5.6 & 9.0 & 64.3 & 44.9 & 20.0 & 8.0 & 18.0 & 14.0 & 15.0 \\
\cmidrule(lr){1-17}
LangMem & 15.3 & 10.1 & 8.0 & 9.6 & 75.1 & 65.6 & 53.6 & 18.5 & 16.3 & 64.0 & 51.2 & 22.0 & 16.0 & 22.0 & 16.0 & 19.0 \\
A-Mem & 14.3 & 8.3 & 7.6 & 9.0 & 64.5 & 43.5 & 41.7 & 14.1 & 20.7 & 55.2 & 43.0 & 18.0 & 12.0 & 18.0 & 14.0 & 15.5 \\
Mem0 & 14.6 & 9.0 & 8.8 & 9.9 & 72.0 & 70.9 & 56.2 & 21.5 & 15.3 & 65.1 & 51.8 & 22.0 & 16.0 & 22.0 & 10.0 & 17.5 \\
Mem0\textsuperscript{g} & 14.4 & 8.8 & 7.7 & 9.1 & 73.9 & 61.4 & 37.4 & 15.6 & 14.6 & 58.7 & 45.8 & 20.0 & 8.0 & 18.0 & 14.0 & 15.0 \\
SimpleMem & 14.8 & 9.3 & 9.1 & 10.2 & 64.8 & 58.7 & 50.2 & 14.6 & 21.2 & 55.7 & 46.3 & 18.0 & 10.0 & 18.0 & 12.0 & 14.5 \\
MemoryOS & 13.6 & 9.8 & 9.0 & 10.0 & 71.0 & 59.7 & 48.5 & 16.2 & 15.4 & 55.7 & 46.8 & 22.0 & 14.0 & 20.0 & 16.0 & 18.0 \\
LightMem & 14.8 & 9.3 & 7.7 & 9.2 & 77.1 & 60.6 & 52.0 & 14.3 & 21.7 & 60.6 & 50.8 & 16.0 & 14.0 & 16.0 & 10.0 & 14.0 \\
\cmidrule(lr){1-17}
\rowcolor{mainmethod}
\textbf{MemPrism}\,\textnormal{\fontsize{8}{9.5}\selectfont (SFT)} & 15.1 & 11.3 & 11.5 & \textbf{12.2} & 69.7 & 78.3 & 42.0 & 30.0 & 18.0 & 48.8 & \textbf{47.8} & 26.0 & 20.0 & 28.0 & 18.0 & \textbf{23.0} \\
\rowcolor{mainmethod}
\textbf{MemPrism}\,\textnormal{\fontsize{8}{9.5}\selectfont (SFT+GRPO)} & 16.3 & 12.3 & 13.4 & \textbf{13.8} & 71.4 & 76.9 & 55.5 & 32.0 & 24.0 & 62.5 & \textbf{54.4} & 30.0 & 22.0 & 32.0 & 20.0 & \textbf{26.0} \\
\arrayrulecolor{black}\bottomrule[0.7pt]
}
\newsavebox{\maintablemethodmeasure}
\sbox{\maintablemethodmeasure}{\textbf{MemPrism}\,\textnormal{\fontsize{8}{9.5}\selectfont (SFT+GRPO)}}
\newlength{\maintablemetricwidth}
\setlength{\maintablemetricwidth}{\dimexpr(\textwidth-\wd\maintablemethodmeasure-34\tabcolsep)/16\relax}
\typeout{Main table equal score-column width: \the\maintablemetricwidth}
\begin{tabular}{l*{16}{w{c}{\maintablemetricwidth}}}
\maintablecontents
\end{tabular}
\end{table*}
\endgroup

\paragraph{Action-conditioned initialization.}
We collect reference trajectories by rolling out a teacher policy on training tasks from each dataset. These trajectories provide historical decision states and their corresponding next actions, but do not directly annotate memory views suitable for those decisions. We therefore use the reference next action to construct initial supervision for view selection. Given a historical decision state and its reference action $u_t^{\mathrm{ref}}$, a view teacher produces a soft preference distribution over candidate specifications:
\begin{equation}
q_T(a\mid t)
=q_T(a\mid o_t,g,\mathcal{H}_{<t},u_t^{\mathrm{ref}}).
\label{eq:view-teacher}
\end{equation}
The reference action allows the view teacher to assess the relative value of presenting different historical relations for the target decision. We initialize the view policy by matching the teacher distribution:
\begin{equation}
\mathcal{L}_{\mathrm{init}}
=\mathbb{E}_t\left[
D_{\mathrm{KL}}\left(
q_T(\cdot\mid t)\,\|\,\pi_\theta^v(\cdot\mid z_t)
\right)\right].
\label{eq:init-loss}
\end{equation}
The reference action is used only to construct offline supervision and is unavailable at inference time.

\paragraph{Outcome-guided optimization.}
The view teacher judges view actions from their specifications, without seeing the rendered views or how the frozen task policy uses them. We therefore refine the view policy with grouped GRPO using downstream task outcomes.

For online interaction tasks, we sample $G$ trajectories for the same task and normalize their within-group returns to obtain trajectory-level advantages:
\begin{equation}
A_i=
\begin{cases}
\dfrac{R_i-\mu_g}{\sigma_g+\epsilon},
&\sigma_g>\epsilon,\\[6pt]
0,&\text{otherwise},
\end{cases}
\label{eq:advantage}
\end{equation}
where $R_i$ is the trajectory return, and $\mu_g$ and $\sigma_g$ are the mean and standard deviation within the group. View decisions in the same trajectory share the advantage $A_i$. We use the clipped GRPO objective, with KL regularization against a fixed reference policy obtained after supervised initialization and an optional entropy bonus. For Mind2Web, action-prediction correctness is the reward for computing within-group relative advantages.

In both training stages, the task policy and view-construction modules remain fixed, and only the view policy is optimized. Training thus directly affects how historical evidence is presented rather than the environment action policy.

\subsection{Relational View Construction}
\label{sec:view-construction}

Given the specification $a_t^v$ selected by the view policy, MemPrism constructs visual working memory for the current decision from the persistent interaction history $\mathcal{H}_{<t}$. This process involves selecting candidate evidence and normalizing events, followed by relational organization and visual rendering.

\paragraph{Why optical rendering?} MemPrism uses optical rendering not merely to compress history, but to expose task-relevant relations in a form that a frozen VLM can directly consume. A structured 2D canvas makes these relations explicit through spatial organization, rather than leaving them to be inferred from a linear text sequence. It also allows symbolic event descriptors to be co-located with visual evidence. Unlike a learned latent-memory representation, the rendered view remains directly inspectable and can be reconstructed from the persistent event stream without introducing a separate learned memory interface.

\paragraph{Candidate evidence and event normalization.}
The constructor first selects historical interactions according to the temporal range $w_t$. Because observations and feedback differ across environments, we use an event extractor $\Phi_b$ adapted to the interaction environment to convert candidate interactions into unified temporary event descriptors:
\begin{equation}
\begin{aligned}
\bar{\mathcal{H}}_t&=\operatorname{Select}_{w_t}(\mathcal{H}_{<t}),\\
D_t&=\Phi_b(\bar{\mathcal{H}}_t),
\end{aligned}
\label{eq:event-normalization}
\end{equation}
where $b$ denotes the interaction environment and $D_t=\{d_i\}$ is the set of extracted event descriptors. Each descriptor is represented as
\begin{equation}
d_i=(i,e_i^{\mathrm{ent}},e_i^{\mathrm{act}},e_i^{\mathrm{out}},\Delta s_i,\xi_i),
\label{eq:event-descriptor}
\end{equation}
recording the event index, main entity, normalized action, observed outcome, sparse state updates, and optional auxiliary evidence, respectively. Event extraction combines model-generated outputs with rule-based validation, determines outcome categories from environmental feedback, and retains only state updates supported by observations. Different environments adapt this shared interface through their respective input-parsing and validation rules: text interactions use action and feedback information, visual interactions additionally incorporate synchronized post-action observations, and web interactions use action metadata and changes in page content before and after the action. This allows subsequent view construction to use a unified event representation without depending on any environment's raw data format.

\paragraph{Relational organization and visual rendering.}
The constructor filters event descriptors according to the outcome condition $c_t$, then organizes and renders the selected evidence as visual working memory through $\operatorname{Render}_{\tau_t,\gamma_t}$:
\begin{equation}
\hat{D}_t=\operatorname{Filter}_{c_t}(D_t),
\qquad
V_t=\operatorname{Render}_{\tau_t,\gamma_t}(\hat{D}_t).
\label{eq:render}
\end{equation}
Given $\hat{D}_t$ and $a_t^v$, the renderer deterministically constructs the selected relational layout. Spatial proximity represents event groups, row and column alignment supports comparisons of action outcomes and state changes, connectors represent state transitions or local event associations, and visual markers highlight exceptions. The granularity $\gamma_t$ controls the number of events displayed and their level of detail. When visual evidence is available, the view may also incorporate synchronized post-action images, validated entity bounding boxes, and object crops.

The event descriptors $D_t$, filtered descriptors $\hat{D}_t$, and visual view $V_t$ are all temporary products of the current decision and are not written back to the persistent history.

\begin{figure*}[!t]
\begin{minipage}[t]{\columnwidth}
\centering
\includegraphics[width=\columnwidth]{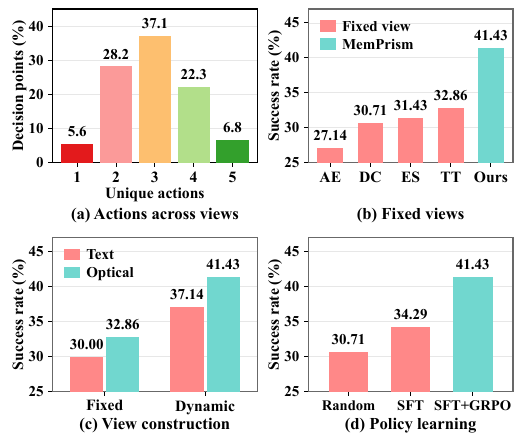}
\caption{Effects of memory views on ALFWorld. (a) Unique actions across five representations of the same history under deterministic decoding. (b) Fixed relation views. (c) View construction and presentation. (d) View policy learning; Random is frequency-matched.}
\Description{Four bar charts in a two-by-two grid. Panel a counts distinct predicted actions at the same decision point under five historical representations: the shares for one through five actions are 5.6, 28.2, 37.1, 22.3, and 6.8 percent. Panel b compares fixed-view success rates, panel c compares organization and presentation, and panel d compares policy learning. Panel a starts at zero percent; the success-rate axes start at 25 percent.}
\label{fig:mechanism}
\end{minipage}\hfill
\begin{minipage}[t]{\columnwidth}
\centering
\includegraphics[width=0.90\columnwidth]{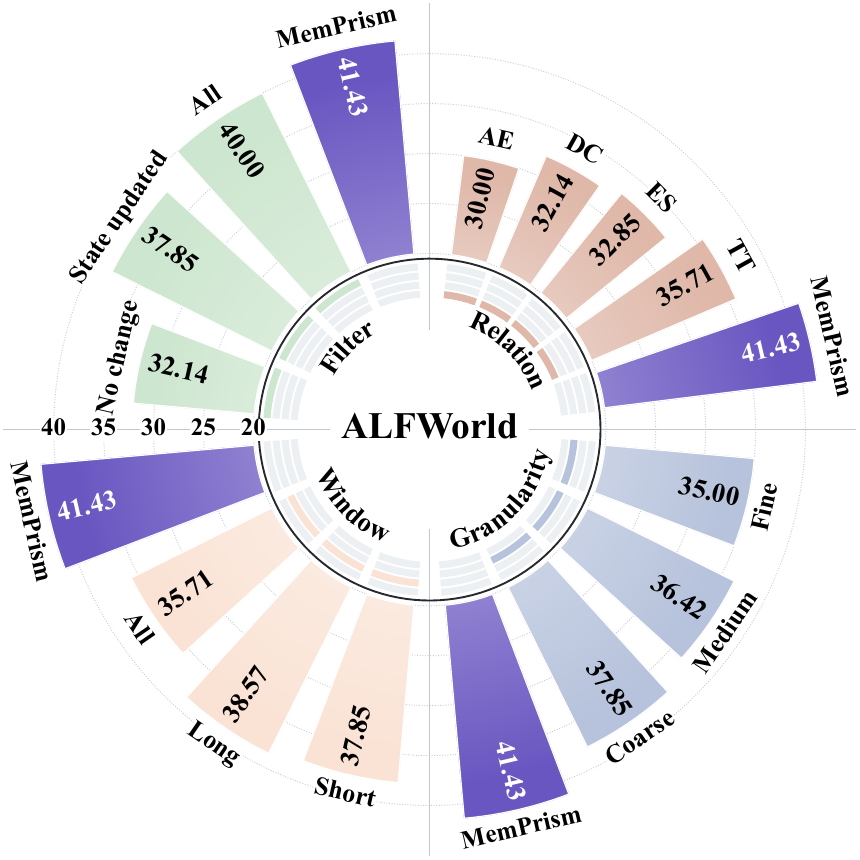}
\caption{Ablation study of the four view-space axes on \mbox{ALFWorld}, measured by success rate (\%).}
\Description{A circular bar chart groups fixed-setting success rates by relation structure, granularity, temporal window, and outcome filter. Four segmented factor rings sit close to the inner circle, with the single-word labels Relation, Granularity, Window, and Filter following their inner edge. ALFWorld appears in the center. A purple MemPrism bar at 41.43 percent appears in each group. Light dotted concentric circles and four radial spokes form the coordinate grid.}
\label{fig:four_axis_ablation}
\end{minipage}
\end{figure*}

\section{Experiments}
\label{sec:experiments}

\subsection{Experimental Setup}
\label{sec:protocol}

\paragraph{Benchmarks}
We evaluate on three benchmarks. \textbf{ALFWorld} \cite{shridhar2020alfred} is a text-based embodied benchmark spanning six household activity categories, with 140 valid-seen games; we report task-weighted success rate (SR). \textbf{EB-ALFRED}, evaluated through EmbodiedBench \cite{yang2025embodiedbench}, measures visual embodied task success; we report SR on Base, Common, Complex, and Long (50 tasks each), and their average. \textbf{Mind2Web} \cite{deng2023mind2web} tests generalization on Cross-Task (252 tasks), Cross-Website (177), and Cross-Domain (912); we report action accuracy. Overall is the task-count-weighted mean across splits. Training-set statistics and evaluation details are in Appendix~\ref{app:training-data}.

\paragraph{Baselines}
We compare against No Memory, AgentOCR (without RL) \cite{feng-etal-2026-agentocr}, LangMem, A-Mem \cite{NEURIPS2025_19909c36}, Mem0 and Mem0$^g$ \cite{chhikara2025mem0}, SimpleMem\cite{liu2026simplemem}, MemoryOS\cite{kang-etal-2025-memory}, and LightMem\cite{fang2026lightmem}. MemPrism-SFT uses supervised view distillation; MemPrism-SFT+GRPO further optimizes the view policy with grouped GRPO. Baseline descriptions and diagnostic variants are detailed in Appendix~\ref{app:baseline-variants}.

\paragraph{Implementation}
Table~\ref{tab:main_results} reports results with frozen Qwen2.5-VL-7B-Instruct and Qwen3.5-4B task policies. By default, analyses use Qwen2.5-VL-7B-Instruct. Phi uses Qwen3.5-2B across all benchmarks. Evaluation uses greedy view selection and task-policy decoding, with step limits of 50 on ALFWorld and 30 on EB-ALFRED. Model configurations are in Appendix~\ref{app:model-config}; training data, rewards, and optimization settings are in Appendix~\ref{app:training-data} and Table~\ref{tab:training-config}.

\begin{figure*}[!t]
\begin{minipage}[t]{\columnwidth}
\centering
\includegraphics[width=0.86\columnwidth]{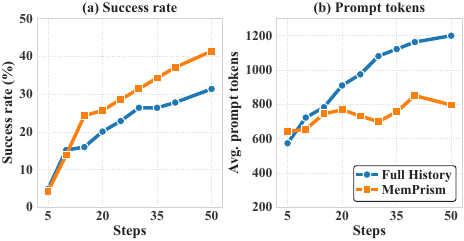}
\caption{ALFWorld success rate (left) and average prompt tokens (right) over cumulative trajectory-length thresholds.}
\Description{Two side-by-side line charts compare Full History and MemPrism over cumulative trajectory-length thresholds: success rate on the left and average prompt tokens on the right.}
\label{fig:long_horizon}
\end{minipage}\hfill
\begin{minipage}[t]{\columnwidth}
\centering
\includegraphics[width=0.86\columnwidth]{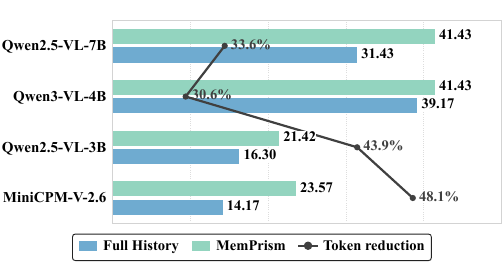}
\caption{Success rates and within-model prompt-token reductions across task VLMs.}
\Description{Horizontal grouped bars compare Full History and MemPrism success rates across four task VLMs. A dark line marks within-model prompt-token reductions. All labels use bold Times New Roman.}
\label{fig:cross_model}
\end{minipage}
\end{figure*}

\subsection{Main Results}
\label{sec:main_results}

Table~\ref{tab:main_results} compares MemPrism with existing memory systems across three benchmarks. On ALFWorld, MemPrism-SFT+GRPO achieves 41.4\% SR, exceeding the strongest baseline, MemoryOS (38.5\%), by 2.9 points. GRPO improves over the SFT-initialized view policy by 7.1 points (34.3\% to 41.4\%). On Mind2Web, MemPrism-SFT+GRPO attains 12.8\% overall action accuracy, compared with 9.4\% for the strongest baseline, LangMem. The gains over the strongest baseline in each split are 1.0 points on Cross-Task, 2.2 on Cross-Website, and 4.2 on Cross-Domain. The largest gain is on Cross-Domain, where MemPrism reaches 12.8\% compared with 8.6\% for AgentOCR. On EB-ALFRED, MemPrism-SFT+GRPO achieves 22.5\% average SR, outperforming the strongest baseline, LangMem (15.5\%), by 7.0 points. The SFT-initialized view policy reaches 19.5\%, exceeding LangMem by 4.0 points; GRPO adds a further \mbox{3.0 points.}

\subsection{Case Study}
\label{sec:case-study}

Figure~\ref{fig:web_case} examines a Mind2Web appointment task at Step 18. A-Mem acts on an outdated Terms state and re-clicks Terms. HiAgent organizes evidence by subgoal, while AgentOCR presents it as a dense chronological image. Although both retain the relevant facts, neither makes the current form state readily accessible for the review decision. These two cases illustrate representation mismatch between the organization of available history and the state integration required by the task. MemPrism selects EntityState to align successive field values, showing that the terms are accepted, SMS updates remain off, and the remaining fields are complete. The task policy then correctly clicks Review Appointment.

\subsection{Effects of Memory Views}
\label{sec:mechanism}

Figure~\ref{fig:mechanism} examines whether view representation changes the frozen policy's next action (a), whether fixed relation structures affect task success (b), and how adaptive construction, presentation medium, and policy learning contribute to performance (c--d).

\textbf{Memory views affect action selection and task success.} At each evaluated ALFWorld decision point, we query the frozen Qwen2.5-VL-7B policy with four relational-view images and an AgentOCR-style image of plain-text history, keeping the decision state, historical evidence, prompt, and remaining view axes fixed under deterministic decoding. Figure~\ref{fig:mechanism}(a) shows that 94.4\% of decision points yield at least two distinct actions, and 66.2\% yield at least three. Since the plain-text image contributes at most one distinct action, the four relational views alone must disagree on at least 66.2\% of decision points. At the task level, we evaluate each relational view under a fully fixed configuration, with the temporal window, granularity, and outcome filter set to recent long, medium, and all, respectively. Success rates range from 27.14\% to 32.86\%, with TemporalTrace performing best---a 5.72-point spread across relation structures (Figure~\ref{fig:mechanism}(b)). Together, these results show that the representation of available history affects both the policy's next action and task-level performance, motivating memory representation as an explicit choice at decision time.

\textbf{Dynamic organization and optical presentation.} Figure~\ref{fig:mechanism}(c) separates adaptive view construction from presentation medium. Fixed Text and Fixed Optical share the TemporalTrace + recent long + medium + all configuration and differ only in presentation medium: 30.00\% vs.\ 32.86\% (+2.86 points). Adaptive construction raises text-based SR to 37.14\% (+7.14) and optical SR to 41.43\% (+8.57). With adaptive construction, optical presentation improves SR by 4.29 points over text. The 8.57-point gain over fixed optical TemporalTrace reflects adaptation across the full view space; the combination of adaptive construction and optical presentation achieves the highest SR.

\textbf{View-policy training improves task success.} Figure~\ref{fig:mechanism}(d) evaluates our training procedure while keeping the task policy and view-construction modules frozen. Frequency-Matched Random samples configurations according to the learned view policy's overall usage frequencies without conditioning on the current context, achieving 30.71\% SR. The SFT-initialized view policy reaches 34.29\%, exceeding this baseline by 3.58 percentage points. Further optimization with grouped GRPO raises SR to 41.43\%, yielding an additional 7.14-point gain over SFT and a total improvement of 10.72 points over Random. These results support the effectiveness of the trained view policy and show that outcome-guided optimization provides further gains \mbox{beyond supervised distillation.}

\subsection{Ablation of the View Space}
\label{sec:view_space_ablation}

Figure~\ref{fig:four_axis_ablation} evaluates the benefit of adapting each view-space axis by fixing one axis at a time while keeping the other three adaptive. MemPrism achieves 41.43\% SR, outperforming the best tested fixed setting on every axis. \textbf{Adapting the relation type yields the largest gain}, improving SR by 5.72 percentage points over the best fixed relation, TemporalTrace (35.71\%). Adapting the temporal window, outcome filter, and granularity yields gains of 2.86, 1.43, and 3.58 points over their respective best fixed settings: recent long (38.57\%), all outcomes (40.00\%), and coarse (37.85\%). These results support adapting how events are related, which events are shown, and how much detail is retained for the current decision.

\subsection{Long-Horizon Behavior}
\label{sec:long_horizon}

Figure~\ref{fig:long_horizon} demonstrates two clear advantages of MemPrism that become pronounced as trajectory length increases.

\textbf{Success rate advantage widens with longer trajectories.}
Full History holds a narrow lead at the shortest thresholds ($\leq$5 and $\leq$10 steps, 0.6--1.3 points), but MemPrism overtakes it at $\leq$15 steps (24.3\% vs.\ 16.0\%) and never trails thereafter. The gap reaches 9.3 points at $\leq$40 steps and 10.0 points at $\leq$50 steps (41.4\% vs.\ 31.4\%). Notably, Full History plateaus after $\leq$30 steps (gaining only 5.0 points from $\leq$30 to $\leq$50), while MemPrism gains an additional 10.0 points over the same span, demonstrating that its advantage compounds as more \mbox{history accumulates.}

\textbf{Prompt tokens stabilize while Full History costs escalate.}
Full History prompt tokens grow nearly linearly with trajectory length, from 574 at $\leq$5 steps to 1,201 at $\leq$50 steps---a 2.1$\times$ increase. In contrast, MemPrism token usage remains tightly bounded between 642 and 852 across all thresholds, with no upward trend beyond $\leq$15 steps. At the $\leq$50-step threshold, MemPrism achieves a 33.6\% token reduction (797 vs.\ 1,201) while delivering a 10.0-point SR improvement, showing that compiled views compress history effectively without losing \mbox{decision-relevant information.}

\subsection{Cross-Model Transfer and Prompt Compression}
\label{sec:cross_model}

Figure~\ref{fig:cross_model} evaluates cross-model transfer using the same view-policy checkpoint without per-model adaptation. Prompt-token \mbox{reductions} are computed within each task VLM; absolute token counts are not compared across models.

\textbf{Universal success rate improvement.} On its training VLM (Qwen2.5-VL-7B), MemPrism raises SR from 31.43\% to 41.43\% (+10.00 pp). The same view policy, applied zero-shot to three unseen VLMs, improves every model: Qwen3-VL-4B from 39.17\% to 41.43\% (+2.26), Qwen2.5-VL-3B from 16.30\% to 21.42\% (+5.12), and MiniCPM-V-2.6 from 14.17\% to 23.57\% (+9.40). The gains are not concentrated at a particular model scale or architecture.

\textbf{Substantial prompt compression across all models.}
Token reduction ranges from 30.64\% (Qwen3-VL-4B) to 48.15\% (MiniCPM-V-2.6), with the training-domain model achieving 33.65\%. In every case, higher SR is achieved with substantially fewer tokens. MiniCPM-V-2.6 exemplifies this interplay: its extreme visual token density (1.8M pixels $\rightarrow$ only 640 tokens) combined with strong OCR makes full-history inputs disproportionately verbose, yielding the largest compression margin (48.15\%) alongside a 9.40-point SR gain. The co-occurrence of SR gains and token reduction across four models indicates that compiled representations are useful across model families.

\section{Conclusion}
We presented MemPrism, a task-conditioned relational memory framework for long-horizon agents. Our central finding is that retrieving the correct evidence is not sufficient: how that evidence is organized determines whether it can support the current decision. Even when the historical facts remain unchanged, different relational views can lead to different outcomes, with dynamic organization and optical presentation contributing complementary benefits. Motivated by this observation, MemPrism decouples stable event storage from the construction of temporary, decision-specific working-memory views. Experiments further show that its advantages become more pronounced as trajectories grow longer,  and that the learned view policy transfers across different VLMs without additional adaptation. Thus, these results identify post-retrieval representation mismatch as a fundamental bottleneck in long-horizon agents and establish task-conditioned memory organization for effective memory use.


\bibliographystyle{ACM-Reference-Format}
\bibliography{references}

\appendix

\section{Reproducibility Statement}
\label{app:reproducibility}

\begin{table*}[t]
\centering
\footnotesize
\caption{Training configurations for the shared view policy.  The architecture and
input tensor shapes are identical across benchmarks; rows below record only
stage-specific optimization settings.}
\label{tab:training-config}
\begin{tabularx}{\textwidth}{@{}p{0.20\textwidth}Y@{}}
\toprule
Setting & Hyperparameters \\
\midrule
D1, shared &
AdamW; learning rate \(10^{-4}\); weight decay \(0.01\); batch size 256;
gradient-norm cap \(1.0\); seed 7. \\
D1, ALFWorld &
50 epochs; cosine schedule; minimum learning-rate ratio \(0.01\);
BF16. \\
D1, ALFRED path &
20 epochs; cosine schedule; minimum learning-rate ratio \(0.01\);
no automatic mixed precision. \\
D1, Mind2Web &
50 epochs; cosine schedule; minimum learning-rate ratio \(0.01\);
BF16. \\
GRPO, ALFWorld/ALFRED shared &
View-policy head in training mode; AdamW; learning rate \(3\times10^{-6}\); zero weight
decay; clipping coefficient \(0.2\); KL coefficient \(0.2\); one update epoch;
update batch size 512; gradient-norm cap \(1.0\); constant learning rate;
rollout temperature \(1.0\). \\
GRPO, ALFWorld &
100 iterations; 8 task groups per iteration; 16 trajectories per group;
maximum 50 environment steps; entropy coefficient \(0.003\). \\
GRPO, ALFRED path &
100 iterations; 8 task groups per iteration; 16 trajectories per group;
maximum 30 environment steps; progress-reward coefficient \(0.2\); entropy
coefficient \(0\). \\
GRPO, Mind2Web &
View-policy head in training mode; learning rate \(10^{-6}\); 8 task groups per iteration;
16 trajectories per group;
\(\epsilon_{\mathrm{low}}=0.1\), \(\epsilon_{\mathrm{high}}=0.15\);
\(\beta_{\mathrm{KL}}=0.01\); rollout temperature \(T=1.0\); top-p \(=0.85\);
reward \(\mathbf{1}[\text{action correct}]\); 150 training steps. \\
\bottomrule
\end{tabularx}
\end{table*}

\subsection{Artifact and result provenance}

The released artifact contains the source code for Recorder, View Policy, Composer,
Renderer, benchmark adapters, training launchers, and evaluation scripts.  Each
reported result is accompanied by a machine-readable run manifest
containing the code revision, working-tree status, data split, ordered episode
list, random seed, model identifiers, checkpoint hashes, prompt version,
decoding settings, and output paths.  Rendered working-memory views are
derived artifacts; the persistent event stream and environment trajectory
remain the sources of truth.

\subsection{Models, checkpoints, and common settings}
\label{app:model-config}

All three benchmarks use the same view-policy architecture and tensor interface.
The main comparisons use Qwen2.5-VL-7B-Instruct and Qwen3.5-4B as frozen task
policies; the textual event extractor (Phi) uses Qwen3.5-2B throughout.
The view policy's text backbone is a frozen \texttt{all-MiniLM-L6-v2}.  The view policy is implemented as
a \texttt{ViewPolicyModel} with \texttt{d\_model} \(=512\), eight attention
heads, four Transformer encoder layers, a 1,024-dimensional feed-forward
sublayer, and dropout \(0.1\).  Final evaluation takes the argmax of the
144-way view-policy distribution and uses greedy task-policy decoding.  The run
manifest binds each reported result to the exact task-policy, Phi, MiniLM, and
view-policy model IDs and SHA-256 checkpoint hashes rather than to local filesystem
paths.

At decision step \(t\), the benchmark adapter serializes the current state as
text \(x_t\) and supplies the task goal \(g\).  MiniLM tokenizes each string to
at most 256 tokens, applies attention-mask-aware mean pooling to the final
hidden states, and \(L_2\)-normalizes the result, producing
\(e_t^o,e^g\in\mathbb{R}^{384}\).

\paragraph{Multi-scale interaction descriptors.}
In Eq.~\eqref{eq:history-statistics},
\(\mathcal{H}_{<t}^{(K)}\) and \(\mathcal{H}_{<t}^{(M)}\) denote the latest
\(K\) and \(M\) interactions, respectively. We set \(K=8\) and \(M=64\).
The local descriptor characterizes recent action repetition, feedback outcomes,
and state-change signals, while the global descriptor summarizes history length,
entity distributions, and outcome statistics over a longer horizon.

\texttt{EventWindowSummarizer} converts the most recent \(K\) raw/\allowbreak structured
event pairs into \(u_t\in\mathbb{R}^{128}\).  Its first eight coordinates
encode the \texttt{action\_frequency} profile used by all three benchmarks.
Raw actions are canonicalized from fields such as \texttt{op},
\texttt{action\_type}, or \texttt{type}, and the eight largest normalized
signature frequencies are stored in descending order.  The next three
coordinates are the exception, state-update, and
no-observed-change ratios.  Seven further coordinates record the normalized
number and coverage of entities, the delta ratio, the maximum exception and
no-change streaks, and recency-weighted state-change and exception densities.
The remaining positions are zero padded to form the fixed 128-dimensional
interface.  This channel summarizes repetition and concentration without
assuming a benchmark-specific structured action taxonomy.  A separate vector
\(s_t\in\mathbb{R}^{8}\), computed over at most the latest \(M\) events,
contains the exception ratio, state-update ratio, normalized unique-entity
count, normalized history length, a deterministic unit hash of the most
recent entity, normalized terminal exception and no-change streaks, and
recent state-update density.

The encoder does not concatenate these heterogeneous features into one flat
input.  Four independent linear layers project \(e_t^o\), \(e^g\), \(u_t\),
and \(s_t\) to 512 dimensions; a learned 512-dimensional \texttt{[CLS]}
token is prepended to form the five-token sequence
\([\mathrm{CLS},o,g,w,s]\).  Sinusoidal positional encodings and dropout are
applied before the four-layer pre-norm Transformer, whose feed-forward blocks
use GELU.  The final \texttt{[CLS]} state passes through a
\(512\!\rightarrow\!512\!\rightarrow\!512\) GELU MLP and LayerNorm to yield
\(z_t\in\mathbb{R}^{512}\).  The current training and evaluation
\texttt{ViewPolicyModel.encode} path does not pass the optional CLIP image
embedding, so the view policy selects views using the five tokens above; environment images
and compiled optical views are consumed downstream by the frozen task policy.
Finally, three residual MLP blocks with 1,024-dimensional hidden layers map
\(z_t\) to a single joint distribution over all 144 view actions.  The
reported \(\tau\), window/filter, and \(\gamma\) marginals are derived from
these joint logits rather than trained by separate heads.

\subsection{Data and evaluation protocols}
\label{app:training-data}

We use only the official training partitions associated with the three
evaluation settings.  Reference trajectories for action-conditioned
initialization are collected by rolling out a teacher policy on tasks from
these training partitions.  ALFWorld distillation and online rollouts draw from its
training games after the launcher removes unsupported task variants.  The
ALFRED path uses the official ALFRED training annotations for learning and
reserves EB-ALFRED episodes for evaluation.  Mind2Web uses its official
training split to collect teacher-generated reference trajectories and
construct action-conditioned view-teacher records.  No evaluation episode is used to
optimize the view policy.

\paragraph{Benchmark splits and metrics.}
ALFWorld provides 3,827 training instances and 140 valid-seen evaluation games
across six household activity categories. We report per-category SR and the
task-weighted average. EB-ALFRED is built on ALFRED and provides RGB and text observations through
EmbodiedBench across six available subsets; our evaluation covers Base, Common,
Complex, and Long, with 50 tasks each. Avg. denotes the average SR over these
four subsets. These results follow the EmbodiedBench protocol rather than the
official ALFRED leaderboard. Mind2Web contains 1,009 training tasks across 137
websites and 31 domains. Its Cross-Task, Cross-Website, and Cross-Domain test
splits contain 252, 177, and 912 tasks, respectively. The model predicts CLICK,
TYPE, or SELECT. We report step-level action accuracy and exact
episode success; Overall is the task-count-weighted mean of the three split
scores.

\paragraph{Baseline and diagnostic variants.}
\label{app:baseline-variants}
LangMem maintains working and long-term memory through episodic
observation--thought--action chains. A-Mem uses Zettelkasten-style atomic notes
with dynamic linking and memory evolution. Mem0 and Mem0$^g$ extract facts via
LLM calls and retrieve top-$k$ similar memories; Mem0$^g$ additionally uses a
Neo4j entity--relation graph. The mechanism analyses include Recent-4 Text,
Full Text History, four fixed relation views (ActionEffect, DependencyChain,
EntityState, and TemporalTrace), dynamically selected text and optical views,
and view policies with one view-space factor fixed. These diagnostic variants
are evaluated separately from the external baselines. MemPrism-SFT uses
supervised view distillation, and MemPrism-SFT+GRPO adds online grouped GRPO
on ALFWorld and the ALFRED training path, or offline replay on Mind2Web.

\paragraph{Cold-start distillation.}
We first train the encoder and joint view policy from the teacher distribution
\(q_t\), before any online policy optimization.  For each decision state, the
loss matches the teacher distribution through
\(\mathrm{KL}(q_t\Vert\pi_\theta)\), as defined in
Equation~\ref{eq:init-loss}.  The 144-way output layer is trained jointly
with the encoder and the three residual MLP blocks of the view-policy head; the unused
compatibility-only entity-hint head remains frozen.  The resulting D1
checkpoint supplies both
the online initialization and the fixed reference policy used by the GRPO KL
term.

\paragraph{Grouped GRPO training.}
For ALFWorld and the ALFRED training path, trajectories for the same task form
a group and their terminal task signals are normalized within that group.
ALFWorld uses binary terminal success as the reward with a 50-step limit.
The ALFRED path uses a 30-step limit and rewards success plus
\(0.2\times\)goal-condition progress. Each iteration uses eight task groups
with 16 trajectories per group (128 trajectories in total).
For Mind2Web, each iteration contains eight offline-replay task groups with
16 trajectories per group, and the binary reward indicates whether the
predicted action is correct.
During this phase, the encoder, Phi, compiler, renderer, frozen task policy,
and D1 reference policy remain fixed.  Optimization updates only the shared
three-block view-policy head and its 144-way joint output layer.

\subsection{Parameter configurations}
\label{app:training-config}

The common architecture is fixed as described in Section~\ref{app:model-config};
Table~\ref{tab:training-config} therefore reports only stage-specific
optimization settings.

\paragraph{Hardware, software, and checkpoints.}
All final runs used one node with eight NVIDIA H200 GPUs; wall-clock training
was 24 h for ALFWorld, 48 h for EB-ALFRED, and 24 h for Mind2Web.  The
peak per-GPU memory footprints were 127.0 GB for ALFWorld, 77.6 GB for
EB-ALFRED, and 127.0 GB for Mind2Web.  ALFWorld and Mind2Web view-policy training
use Python 3.10 and Transformers 4.51.1 with a CUDA-matched PyTorch,
and ALFWorld model serving uses vLLM 0.8.5--0.11.0.  EB-ALFRED uses
Python 3.9.20, Torch 2.4.0, AI2-THOR 2.1.0, NumPy 1.26.4, and Transformers
4.57.6 for simulation and view-policy-head updates, together with Python 3.12, Torch
2.11.0, SGLang 0.5.14, Transformers 5.8.1, and NVIDIA driver 580.126.20 for
model serving.  Epoch and GRPO checkpoints store the full model, encoder,
view-policy head, optimizer state, stage, epoch or iteration, metrics, and
configuration; exported reference or final weight-only checkpoints may omit
optimizer state by design.

\section{EB-ALFRED Case Study}
\label{app:case-studies}

The following case study presents a decision in prompt--response order.  It
includes the decision-relevant prompt excerpt, the two visual inputs delivered
to the frozen task policy, the logged model output, and the resulting progress
update.  The implementation
concatenates the shared policy prefix, decision-specific suffix, current
camera image, and compiled view into one multimodal \texttt{role=user}
message.

\refstepcounter{subsection}
\label{app:case-entity-state}

\begin{figure*}[p!]
\centering
\begin{caseflowbox}{EntityState Case Study}
\begin{casepromptbox}{\casebadge{caseblue}{SYSTEM} Policy prompt}
\small
\textbf{Task.} Move the two books from the bed to the desk.\\[2pt]
The EB-ALFRED task-policy prompt requires
reasoning inside \texttt{<think>} tags followed by exactly one admissible
action inside \texttt{<action>} tags.
\end{casepromptbox}

\casearrow

\begin{casepromptbox}{\casebadge{caseblue}{USER} Decision-specific prompt}
\small
\textbf{Current state.} At Step 21, the desk is visible and the robot is
holding a book.\\[2pt]
\textbf{Recent text history.} Step 18 successfully found a book; Step 19
successfully found the desk.\\[2pt]
\textbf{Attachments.} The first image is the current camera observation; the
second is the compiled \texttt{EntityState} view.
\end{casepromptbox}

\casearrow

\begin{casevisualbox}{\casebadge{caseblue}{VISION} Visual inputs to the frozen task policy}
\begin{minipage}[c]{0.39\linewidth}
  \centering
  \includegraphics[width=\linewidth]{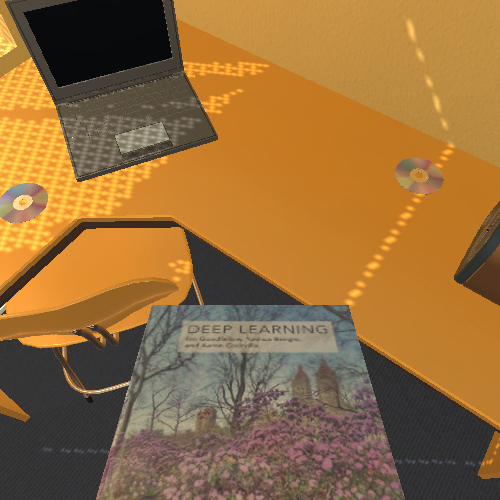}\\[2pt]
  \footnotesize Current observation
\end{minipage}\hfill
\begin{minipage}[c]{0.56\linewidth}
  \centering
  \includegraphics[width=\linewidth]{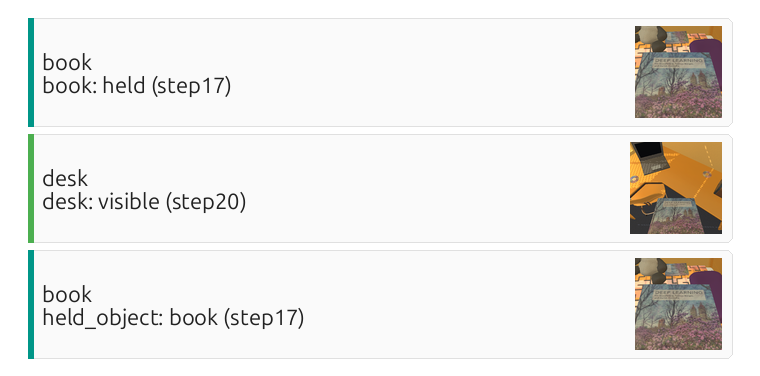}\\[2pt]
  \footnotesize Compiled \texttt{EntityState}
\end{minipage}
\end{casevisualbox}

\casearrow

\begin{caseresponsebox}{\casebadge{casegreenline!70!black}{VLM} Assistant output}
\small
\texttt{<think>} I need to place the book on the desk as part of the task.  The
book is currently being held, and a visible desk has been located.  The next
step is to put down the book on the desk. \texttt{</think>}\\[2pt]
\texttt{<action>put down the object in hand</action>}
\end{caseresponsebox}

\casearrow

\begin{caseresultbox}{\casebadge{caseblue}{RESULT} Recorded outcome}
\small
Last action executed successfully.  \textbf{Progress: $0\rightarrow1/2$.}
State updates: \texttt{book -> on desk}; \texttt{held\_object -> none}.
\end{caseresultbox}
\end{caseflowbox}
\caption{Case study of \method{} on EB-ALFRED.}
\Description{EB-ALFRED entity-state case study showing the current observation, relational memory view, policy response, and recorded outcome.}
\label{fig:case-entity-state}
\end{figure*}

\section{Optical-Resolution Sensitivity}
\label{app:resolution}

We evaluate the final \method\ checkpoint on ALFWorld with maximum image long
sides of 448, 672, and 896 pixels.  The checkpoint, compiler, rendering rules,
prompt template, ordered episode list, and decoding settings are fixed; the
maximum-long-side cap is the only configured factor.  Online trajectories,
including later view-policy inputs and selected views, are allowed to diverge
naturally after an action differs.

\begin{table}[H]
\centering
\footnotesize
\caption{ALFWorld resolution sensitivity under three maximum-long-side
limits.}
\label{tab:resolution}
\begin{tabular}{@{}rrrr@{}}
\toprule
Max side (px) & SR (\%) & Avg. prompt tok. & Invalid action (\%) \\
\midrule
448 & 38.57 & 703 & 0.043 \\
672 & \textbf{41.43} & 797 & 0.064 \\
896 & 40.00 & 917 & 0.097 \\
\bottomrule
\end{tabular}
\end{table}

The 672-pixel setting obtains the highest SR at \(41.43\%\), with 797 average
prompt tokens and a \(0.064\%\) invalid-action rate.  Reducing the longest side
to 448 pixels lowers the average prompt length to 703 tokens and the
invalid-action rate to \(0.043\%\), while SR decreases to \(38.57\%\).
Increasing the longest side to 896 pixels raises the average prompt length to
917 tokens and the invalid-action rate to \(0.097\%\), without improving over
the 672-pixel SR.  We therefore retain 672 pixels as the default setting.

\end{document}